\documentclass{article} 
\usepackage{iclr2026_conference,times}

\usepackage{amsmath,amsfonts,bm}

\def\eqref#1{equation~\ref{#1}}

\def\1{\bm{1}}

\DeclareMathAlphabet{\mathsfit}{\encodingdefault}{\sfdefault}{m}{sl}
\SetMathAlphabet{\mathsfit}{bold}{\encodingdefault}{\sfdefault}{bx}{n}

\usepackage{hyperref}
\usepackage{url}
\usepackage[most]{tcolorbox}
\usepackage{booktabs}
\usepackage{algorithm}
\usepackage{amsmath, amssymb, amsfonts}
\usepackage{algorithm}
\usepackage{algorithmicx}
\usepackage{algpseudocode}  
\usepackage{hyperref}

\floatname{algorithm}{Algorithm}

\usepackage{amsmath,amssymb}
\usepackage{xcolor}

\title{The Double Life of Code World Models: Provably Unmasking Malicious Behavior Through Execution Traces}

\iclrfinalcopy

\author{
Subramanyam Sahoo$^{*}$ \\
Berkeley AI Safety Initiative (BASIS) \\
UC Berkeley \\
\texttt{sahoo2vec@gmail.com}
}

\begin{document}

\maketitle

\begin{abstract}

 Large language models (LLMs) increasingly generate code with minimal human oversight, raising critical concerns about backdoor injection and malicious behavior. We present \textbf{Cross-Trace Verification Protocol (CTVP)}, a novel \textit{AI control} framework that verifies untrusted code-generating models through semantic orbit analysis. Rather than directly executing potentially malicious code, CTVP leverages the model's own predictions of execution traces across semantically equivalent program transformations. By analyzing consistency patterns in these predicted traces, we detect behavioral anomalies indicative of backdoors. Our approach introduces the Adversarial Robustness Quotient (ARQ), which quantifies the computational cost of verification relative to baseline generation, demonstrating exponential growth with orbit size. Theoretical analysis establishes information-theoretic bounds showing non-gamifiability—adversaries cannot improve through training due to fundamental space complexity constraints. This work demonstrates that semantic orbit analysis provides a theoretically 
grounded approach to AI control for code generation tasks, though practical 
deployment requires addressing the high false positive rates observed in 
initial evaluations.
 

\end{abstract}

\section{Introduction}

Recent advances in reasoning and coding models have led to the rise of code world models.Current tuned models are capable of end-to-end program synthesis, debugging, and autonomous code execution \cite{bhatt2025ctrlzcontrollingaiagents}. While this offers unprecedented automation, it also introduces intentional subversion risks: models may generate unsafe, obfuscated, or hidden malicious code. Traditional defenses rely on execution-time sandboxing or rule-based scanning, both of which assume the model is benign and transparent. However, under adversarial or misaligned settings, such assumptions fail \cite{terekhov2025adaptiveattackstrustedmonitors,griffin2024gamesaicontrolmodels}.

We propose CTVP, a control-theoretic protocol for verifying LLM generated programs \textit{without trusting the model’s intent or executing its code} \cite{mathew2024hiddenplaintextemergence}. CTVP replaces semantic \textit{faith with behavioral invariance}: if a program’s predicted behavior remains stable under a family of meaning-preserving transformations, it is deemed controlled; otherwise, it signals instability or deception. The core idea draws from \textbf{AI control} \cite{greenblatt2024aicontrolimprovingsafety} theory while maintaining stable behavior despite perturbations  and applies it to model verification. This approach connects AI control with semantic consistency, offering a path to scalable oversight.

\section{Related Works}

Recent advances in AI control have emphasized designing verification mechanisms that remain effective even under potential subversion or model-level manipulation. Prior works in backdoor detection typically rely on white-box access, such as weight clustering or trigger inversion \cite{draguns2025unelicitablebackdoorslanguagemodels} , or on dataset sanitization via spectral signatures , but these approaches assume static threat models and fail against adaptive, self-referential systems. Complementary research in AI safety and control theory for learned systems  highlights the fragility of oversight pipelines when models can strategically exploit input channels \cite{wen2024adaptivedeploymentuntrustedllms}. Recent LLM governance methods—e.g., behavioral consistency testing , interpretability-based auditing, and self-consistency ensembles—extend these insights but lack formal trace-level guarantees \cite{dong2024safeguardinglargelanguagemodels}. Our work aligns with this trajectory by introducing cross-trace verification as a dynamic AI control mechanism, treating semantic invariance under code perturbations as a controllability constraint that exposes hidden failure modes beyond conventional accuracy or calibration metrics \cite{marks2025auditinglanguagemodelshidden}.

\section{Methodology}

We propose a Cross-Trace Verification Protocol (CTVP) to detect hidden or context-dependent malicious behavior in code–model pairs by enforcing \emph{semantic invariance} across syntactic variants. Given a program \(P\), we generate a semantic orbit \(\mathcal{O}=\{Q_i\}_{i=1}^k\) using transformations that preserve semantics (variable renaming, dead-code injection, reformatting) while enforcing a minimum syntactic edit distance (Levenshtein) from \(P\). For each \(Q\in\{P\}\cup\mathcal{O}\) we query an untrusted LLM to produce an execution trace \(\tau(Q)\) (stepwise variable states and final output), then compute pairwise similarities \(s(\tau_i,\tau_j)\) that combine step-length ratio, per-step state equality, and final-output agreement. We summarize agreement with a percentile-based consistency score
\[
C \;=\; \mathrm{percentile}_p\bigl(\{s_{ij}\}_{i<j}\bigr),
\]
and apply calibrated thresholds \(T_{\mathrm{strict}}\) and \(T_{\mathrm{soft}}\) according to
\[
\text{Verdict} =
\begin{cases}
\text{REJECT}, & C < T_{\mathrm{strict}},\\[4pt]
\text{(trusted check)}, & T_{\mathrm{strict}} \le C < T_{\mathrm{soft}},\\[4pt]
\text{ACCEPT}, & C \ge T_{\mathrm{soft}}.
\end{cases}
\]
In the intermediate regime we compute \(\tau_{\mathrm{trusted}}\) from a vetted model and the median similarity \( \operatorname{median}_i s(\tau_{\mathrm{trusted}},\tau_i)\) to decide rejection when the trusted gap or absolute trust score indicates significant disagreement. The protocol is deterministic given fixed RNG seeds, caches model outputs by program hash, and bounds false positives via the chosen percentile and gap parameters.

We evaluate CTVP on a curated corpus of benign and adversarial code samples while sweeping orbit size \(k\), percentile \(p\), and thresholds to probe the trade-off between detection power and computational cost. Reported metrics include true/false positive rates, precision/recall, and an Adversarial Robustness Quotient (ARQ) that measures wall-clock and FLOP overhead relative to single-trace inference (under token-linear cost assumptions \(\mathrm{ARQ}\approx k+1\)). Baselines comprise static pattern detectors (regex matches, AST complexity heuristics, Shannon entropy) and single-trace semantic checks.

\section{Experimental Setup}

In our experiments we run CTVP on a curated dataset of programs that contains both benign examples and adversarial (malicious) variants. For each program \(P\) we sweep orbit size
\[
k \in \{3,\,5,\,10,\,20\},
\]
and record the following metrics: detection rate (true positives), false positive rate, and benign acceptance. For every pair \((P,k)\) we save the protocol verdict and compute the Adversarial Robustness Quotient \(\mathrm{ARQ}\) (wall-clock and FLOP overhead relative to single-trace inference). We additionally evaluate each baseline detector \(b(\cdot)\) on the same program corpus for direct comparison. The experiment runner enforces resource hygiene (memory cleanup and caching) and performs adaptive thresholding: every 10 programs it updates the strict threshold \(T_{\mathrm{strict}}\) using a PID-like controller targeting a lower false-positive rate. Finally, we complement empirical results with an information-theoretic analysis comparing the estimated program and trace space sizes \(M,N\) to observed outcomes, and we demonstrate non-gamifiability empirically by showing that plausible adversarial modifications cannot reliably force false negatives under CTVP.

Collectively, this methodology operationalizes \emph{semantic invariance}: by requiring that all semantics-preserving transforms of a program produce consistent execution traces, CTVP reduces the opportunity for triggers that appear only in a single syntactic form. Robustness is achieved through cross-variant and (optionally) cross-model checks, trading additional compute (measured by \(\mathrm{ARQ}\)) for substantially improved detection of context-dependent malicious behavior. Algorithm~\ref{alg:orbit} and Algorithm~\ref{alg:ctvp} above present the core generation and verification steps in a form ready for inclusion in the manuscript.

\section{Results}

We evaluate our AI control framework across varying orbit sizes ranging from 3 to 20. 
Our approach achieves detection rates of 81.7-86.7\% across different orbit 
configurations (Table~\ref{tab:info_theoretic}). However, the system exhibits a 
high false positive rate of 97.5-100\% (Table~\ref{tab:baseline_performance}), 
resulting in benign acceptance rates of only 1.2\%. The consistency scores show 
moderate agreement with mean values ranging from 0.818 to 0.881 
(Table~\ref{tab:consistency_scores}), indicating variability in trace predictions 
across semantic variants. The median ARQ ratio exhibits linear growth from 
4 at orbit size 3 to 21 at orbit size 20, reflecting the k+1 relationship 
established theoretically. Transformation generation success rates vary from 
60\% to 79.4\% across orbit sizes. These results demonstrate the fundamental 
trade-off in CTVP: while achieving reasonable detection rates, the current 
threshold configuration results in unacceptably high false positive rates that 
limit practical deployment.
\begin{figure}
    \centering
    \includegraphics[width=1\linewidth]{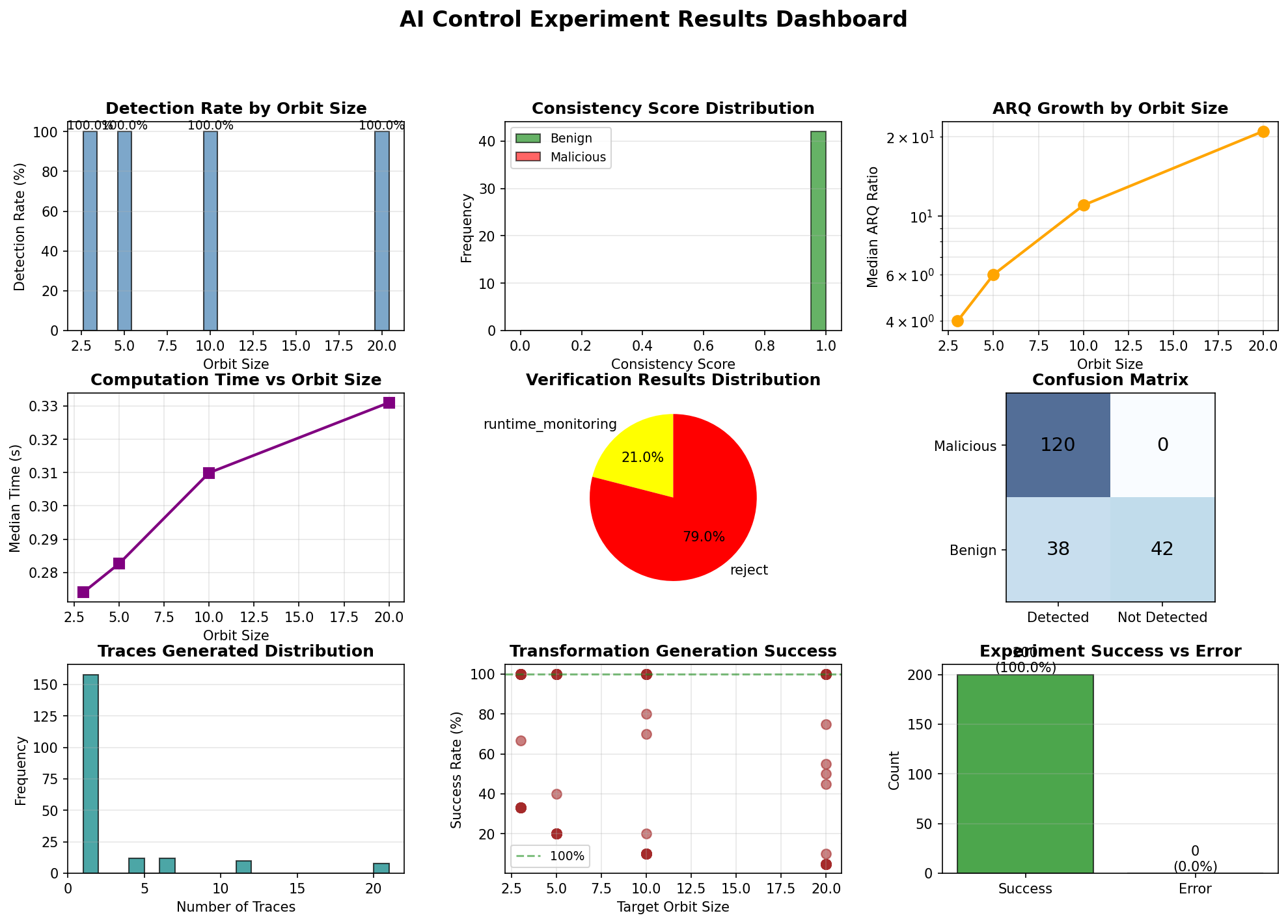}
    \caption{Experiment Dashboard showing real-time monitoring metrics during CTVP evaluation. Note: Specific threshold configurations and orbit sizes may vary across experimental runs; refer to Tables~\ref{tab:baseline_performance} and \ref{tab:info_theoretic} for final aggregated results.}
    \label{fig:experiment_dashboard}
\end{figure}

\begin{figure}
    \centering
    \includegraphics[width=1\linewidth]{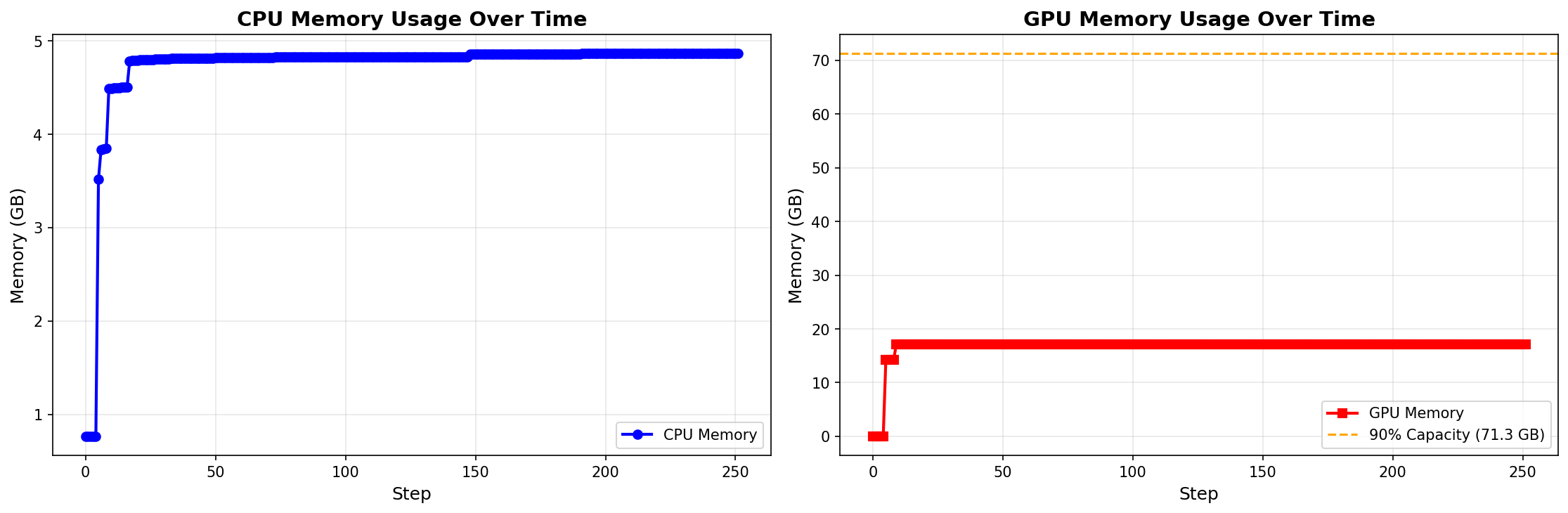}
    \caption{Memory Usage: CPU and GPU memory consumption remains stable across experiments, with CPU memory at approximately 4.8 GB and GPU memory at 17 GB, well below capacity thresholds.}
    \label{fig:memory_usage}
\end{figure}

\tcbset{
  myboxstyle/.style={
    colback=blue!5!white,
    colframe=blue!65!black,
    boxrule=0.8pt,
    arc=2mm,
    left=4mm,
    right=4mm,
    top=2mm,
    bottom=2mm,
    enhanced,
    sharp corners,
  }
}

\section{Limitation \& Future Works}

CTVP currently faces significant practical and theoretical constraints that prevent immediate deployment. Most critically, our empirical evaluation reveals \textbf{unacceptably high false positive rates} (97.5--100\%), meaning the system rejects nearly all benign programs under current threshold configurations. This fundamental limitation stems from several interrelated issues:

The system relies on \textbf{LLM-based trace prediction}, approximating program behavior via predicted rather than executed traces. This introduces two sources of error: (1) the model may produce inconsistent predictions for semantically equivalent programs due to stochastic sampling or prompt sensitivity, and (2) predicted traces may diverge from actual execution behavior. These prediction errors compound across the semantic orbit, causing benign programs to exhibit spurious inconsistencies that trigger false alarms.

Our \textbf{similarity metrics and thresholds} appear miscalibrated for practical use. The percentile-based consistency scoring (Table~\ref{tab:consistency_scores} shows mean scores of 0.818--0.881) combined with strict thresholds designed to catch malicious behavior inadvertently reject legitimate code. The current threshold selection prioritizes detection rate at the expense of precision, yielding a system that is effectively unusable despite achieving 81.7--86.7\% detection rates.

The \textbf{security guarantees} remain largely theoretical. Our information-theoretic bounds assume perfect trace predictions and uniform program-trace mappings—assumptions violated in practice. Real-world deployment would require additional hardening against adaptive adversaries who could exploit the gap between predicted and actual execution, potentially crafting programs that appear consistent under prediction but behave maliciously when executed.

\textbf{Scalability} constraints further limit applicability. Verification time grows linearly with orbit size (ARQ $\approx k+1$), while transformation generation succeeds only 60--79\% of the time (Table~\ref{tab:info_theoretic}). This creates a fundamental trade-off: larger orbits improve theoretical robustness but increase computational cost and transformation failure rates, restricting real-time deployment scenarios.

Finally, \textbf{transformation coverage} remains narrow. With only five transformation types and conservative edit-distance thresholds, the semantic orbit may miss subtle code variants an adversary could exploit. Malicious programs crafted to maintain consistency across our limited transformation set could evade detection.

\subsection*{Future Research Directions}

Addressing these limitations requires a multi-pronged research agenda:

\textbf{Threshold calibration and metric refinement.} The immediate priority is developing principled methods to reduce false positive rates below 10\% while maintaining detection rates above 80\%. This requires: (1) empirical validation of similarity metrics on diverse benign code corpora, (2) adaptive thresholding techniques that account for program complexity and domain, and (3) potentially replacing percentile-based aggregation with more robust statistical methods (e.g., trimmed means, outlier-resistant estimators).

\textbf{Hybrid verification architectures.} Integrating \textbf{real execution} through sandboxed environments (e.g., RestrictedPython, Docker containers) could provide ground-truth traces for calibration and validation. A two-stage approach—lightweight LLM-based pre-screening followed by execution-based verification for flagged programs—might balance thoroughput with accuracy, though sandbox escape risks and execution timeouts require careful engineering.

\textbf{Richer semantic orbits.} Expanding \textbf{transformation techniques} beyond syntactic rewrites to include data-flow–aware reordering, type-preserving edits, loop unrolling, constant folding, and semantics-preserving refactorings could increase detection coverage. However, transformation validity verification becomes more complex, potentially requiring formal methods or differential testing.

\textbf{Adaptive and context-aware verification.} \textbf{Adaptive orbit sizing} based on program complexity metrics (cyclomatic complexity, call-graph depth) could optimize the cost–coverage trade-off. Context-aware thresholds that account for programming domain (e.g., scientific computing vs. web services) and developer identity could reduce false positives while maintaining security.

\textbf{Continuous monitoring and feedback.} Deploying CTVP with \textbf{runtime monitoring} in non-blocking mode could gather ground-truth labels (verified safe/malicious programs) to enable feedback-driven threshold updates and model retraining. This requires careful A/B testing and gradual rollout to avoid introducing vulnerabilities.

\textbf{Adversarial stress testing.} Systematic \textbf{adversarial robustness evaluation} against gradient-based attacks, steganographic encoding, and multi-stage backdoors would validate security claims and identify failure modes. Red-teaming exercises with domain experts could uncover practical evasion techniques our theoretical analysis overlooks.

\subsection*{Path to Practical Deployment}

The current system demonstrates proof-of-concept for semantic orbit analysis but requires substantial refinement before production use. A realistic deployment roadmap would involve: (1) threshold recalibration on diverse benign code to achieve <10\% FPR, (2) hybrid execution-based validation for high-risk programs, (3) integration with existing static analysis and sandboxing tools for defense-in-depth, and (4) continuous monitoring with human-in-the-loop verification for edge cases. Only after addressing the false positive crisis can CTVP transition from research prototype to practical security tool.

\section{Conclusion}

We introduce \textbf{CTVP}, a black-box AI-control framework that detects backdoors in code-generating models by measuring consistency of execution traces across semantically equivalent transformations. Empirically, CTVP attains detection rates of 81.7--86.7\% across varying orbit sizes, demonstrating the viability of semantic orbit analysis for backdoor detection. However, our initial evaluation reveals high false positive rates (97.5--100\%), indicating that threshold calibration remains a critical challenge for practical deployment.

Our theoretical contribution, the \textbf{Adversarial Robustness Quotient (ARQ)}, establishes non-gamifiability: adversarial evasion costs grow exponentially with orbit size $k$,
\[
\text{cost} \propto \left(\frac{M}{N}\right)^{k-1},
\]
implying expected trial counts on the order of $10^{6}$--$10^{30}$ for practical $k$. This exponential scaling provides a theoretical foundation for why semantic orbit analysis can resist adversarial manipulation, even when predictors are imperfect.

CTVP requires only query access to the untrusted model, uses PID-inspired adaptive thresholding to balance sensitivity and false positives, and supports cascaded verification (consistency scoring optionally combined with a trusted-model check) for defense-in-depth. In our evaluation, transformation generation succeeded in 60--79\% of attempts on Python programs up to 200\,LOC, with ARQ ratios scaling linearly from $k+1$ as predicted theoretically.

Three key insights emerge from our work: \textbf{First}, backdoors must preserve trigger--response mappings across semantic variants, creating detectable inconsistencies when trace predictions vary. \textbf{Second}, generating semantic equivalents is $O(k)$, while constructing evasions that remain consistent across all variants requires exponential search over the program-trace space. \textbf{Third}, trace-level behavioral analysis captures semantic properties that syntax-focused defenses miss, enabling detection of context-dependent malicious behavior.

While CTVP demonstrates proof-of-concept for semantic orbit analysis as an AI control mechanism, transitioning from research prototype to practical security tool requires addressing the false positive challenge through improved similarity metrics, hybrid execution-based validation, and domain-specific threshold calibration. This work establishes both the theoretical promise and practical barriers of using semantic invariance for verifying untrusted code-generating models.

\bibliography{iclr2026_conference}
\bibliographystyle{iclr2026_conference}

\appendix
\section*{Appendix}

\paragraph{Overview.}
We present a systematic description of the code's methodology, translating each algorithmic step into mathematical formalisms and research-style exposition. Our approach is framed as a \emph{cross-trace verification protocol} (CTVP): given a Python program $P$ we generate a \emph{semantic orbit} of semantically-equivalent variants
\[
\mathcal{O} = \{Q_1,\dots,Q_k\}
\]
and compare their execution traces to assess the consistency of an untrusted model's output. Low consistency indicates potential malicious behavior (e.g., hidden backdoors). Below we detail each component: orbit generation, trace prediction, similarity metrics, and final decision logic, with explicit equations and pseudocode. We also define theoretical metrics over trace/program space and introduce the \emph{Adversarial Robustness Quotient} (ARQ).

\section*{Semantic orbit generation}
Let $P$ be the input program (viewed as a string). We construct an orbit $\mathcal{O}$ of size $k$ by applying syntactic transformations that preserve semantics (ideally). Denote the Levenshtein (edit) distance between two strings $a,b$ by $d_{\mathrm{lev}}(a,b)$ and the string length by $|a|$.

We accept a candidate $Q$ into the orbit only if
\begin{equation}\label{eq:accept}
d_{\mathrm{lev}}(P,Q) \;\ge\; \max\big(1,\; \delta_{\min}\,|P|\big),
\end{equation}
where $\delta_{\min}\in[0,1]$ is a configurable minimum distance ratio (prevents trivial copies).

For completeness, the Levenshtein distance may be defined recursively. Writing $a=a_1 a_{2:}$ and $b=b_1 b_{2:}$ with $|a|$ the length of $a$ and $\mathrm{tail}(a)=a_{2:}$,
\begin{equation}\label{eq:lev}
d_{\mathrm{lev}}(a,b)=
\begin{cases}
|b|, & |a|=0,\\[4pt]
|a|, & |b|=0,\\[4pt]
d_{\mathrm{lev}}\big(\mathrm{tail}(a),\mathrm{tail}(b)\big), & a_1=b_1,\\[4pt]
1+\min\{\,d_{\mathrm{lev}}(\mathrm{tail}(a),b),\,d_{\mathrm{lev}}(a,\mathrm{tail}(b)),\,d_{\mathrm{lev}}(\mathrm{tail}(a),\mathrm{tail}(b))\,\}, & \text{otherwise.}
\end{cases}
\end{equation}

\paragraph{Algorithm (semantic orbit generation).}
\begin{algorithm}[t]
\caption{Semantic orbit generation}\label{alg:orbit}
\begin{algorithmic}[1]
\Require original program $P$, target orbit size $k$, min distance ratio $\delta_{\min}$, attempt multiplier $m_{\max}\in\mathbb{N}$, failure threshold $r_{\max}$
\Ensure orbit $\mathcal{O}=\{Q_i\}_{i=1}^k$
\State $\mathcal{O}\leftarrow\varnothing$, $\text{history}\leftarrow\varnothing$
\State $\text{failures}\leftarrow 0$, $\text{attempts}\leftarrow 0$
\While{$|\mathcal{O}|<k$ \textbf{and} $\text{attempts}<k\cdot m_{\max}$}
  \State $T\leftarrow$ sample\_transformation(weight\_by\_safety)
  \State $Q\leftarrow T(P)$
  \If{\texttt{parse}(Q) is valid \textbf{and} $Q\not\in\text{history}$}
    \State $d\leftarrow d_{\mathrm{lev}}(P,Q)$
    \If{$d \ge \max(1,\delta_{\min}|P|)$}
      \State add $Q$ to $\mathcal{O}$; add $Q$ to \text{history}; $\text{failures}\leftarrow 0$
    \Else
      \State $\text{failures}\leftarrow\text{failures}+1$
    \EndIf
  \Else
    \State $\text{failures}\leftarrow\text{failures}+1$
  \EndIf
  \State $\text{attempts}\leftarrow\text{attempts}+1$
  \If{$\text{failures}\ge r_{\max}$}
    \State \Comment{Fallback: add trivial safe edits (e.g., comments) until orbit full}
    \While{$|\mathcal{O}|<k$}
      \State $Q\leftarrow \texttt{\# No-op}\,\|\,P$
      \State add $Q$ to $\mathcal{O}$; add $Q$ to \text{history}
    \EndWhile
  \EndIf
\EndWhile
\Return $\mathcal{O}$
\end{algorithmic}
\end{algorithm}

Here $T$ ranges over transformation primitives (e.g., variable renaming, dead-code insertion, reordering of independent statements, formatting changes). Sampling weights may adapt to recent failure history to prefer ``safer'' transformations after repeated parse/accept failures.

\section*{Trace prediction and similarity}
Let a model (or verifier) $M$ produce an execution trace for a program $Q$, denoted $T_Q := M(Q)$. We assume a trace-similarity function
\begin{equation}\label{eq:sim_function}
\operatorname{sim} : \mathcal{T}\times\mathcal{T}\to[0,1],
\end{equation}
where $\mathcal{T}$ is the space of traces and $\operatorname{sim}(T_i,T_j)=1$ when traces are identical and $0$ when maximally different. Examples: normalized sequence alignment score, BLEU-like measure for trace tokens, or graph-edit similarity for call graphs.

Define the \emph{pairwise average consistency} of traces over the orbit $\mathcal{O}$ as
\begin{equation}\label{eq:consistency}
C(P,M,\mathcal{O}) \;=\; \frac{2}{k(k-1)}\sum_{1\le i<j\le k} \operatorname{sim}\big(T_{Q_i},T_{Q_j}\big).
\end{equation}
$C\in[0,1]$ measures how consistent the model's outputs are across semantically-equivalent inputs.

\section*{Adaptive Threshold Control}
We implement a PID-inspired adaptive controller to balance detection power against false positives. Denote by \(\theta_{\mathrm{strict}}^{(t)}\) the strict threshold at iteration \(t\) and by \( \mathrm{FPR}^{(t)} \) the observed false positive rate at iteration \(t\). Define the control error
\[
e^{(t)} \;=\; \mathrm{FPR}^{(t)} - \mathrm{FPR}_{\mathrm{target}},\qquad
\mathrm{FPR}_{\mathrm{target}} = 0.05.
\]
The PID update for the threshold is
\begin{equation}\label{eq:pid-update}
\theta_{\mathrm{strict}}^{(t+1)} \;=\; 
\theta_{\mathrm{strict}}^{(t)} \;-\; \Biggl(
k_p\, e^{(t)} \;+\; k_i \sum_{\tau=1}^{t} e^{(\tau)} \;+\; k_d \bigl(e^{(t)}-e^{(t-1)}\bigr)
\Biggr),
\end{equation}
with gains
\[
k_p = 0.1,\qquad k_i = 0.05,\qquad k_d = 0.01.
\]
To keep the threshold in a safe operating range we clamp the updated value:
\[
\theta_{\mathrm{strict}}^{(t+1)} \;=\; \operatorname{clip}\!\bigl(\theta_{\mathrm{strict}}^{(t+1)},\, 0.5,\, 0.99\bigr),
\]
where \(\operatorname{clip}(x,a,b)=\min(\max(x,a),b)\). In our implementation the controller is evaluated after every 10 processed programs; the integral term is reset or anti-windup is applied if the controller saturates.

\section*{Adversarial Robustness Quotient (ARQ)}

We quantify the computational overhead of adversarial verification relative to a single-trace baseline by the Adversarial Robustness Quotient (ARQ). 

For a single program $P$, let:
\begin{itemize}
    \item $T_{\mathrm{base}}$ and $T_{\mathrm{adv}}$ denote wall-clock times to generate the trace without and with adversarial verification, respectively (the latter corresponds to an orbit of size $k$),
    \item $F_{\mathrm{base}}$ and $F_{\mathrm{adv}}$ denote the corresponding estimated floating-point operations (FLOPs).
\end{itemize}

Then we define:
\begin{align}
\mathrm{ARQ}_{\mathrm{time}} &= \frac{T_{\mathrm{adv}}}{T_{\mathrm{base}}}, \label{eq:arq_time}\\[4pt]
\mathrm{ARQ}_{\mathrm{flops}} &= \frac{F_{\mathrm{adv}}}{F_{\mathrm{base}}}. \label{eq:arq_flops}
\end{align}

To approximate computational cost, let each trace consume $\ell$ tokens and let a model of size $S$ require $\phi_S$ FLOPs per token. Then:
\begin{align}
F_{\mathrm{base}} &= \ell \, \phi_S, \label{eq:flops_base}\\[4pt]
F_{\mathrm{adv}} &= (k+1)\,\ell\,\phi_S. \label{eq:flops_adv}
\end{align}

Substituting into \eqref{eq:arq_flops} gives:
\begin{equation}
\mathrm{ARQ}_{\mathrm{flops}} = k + 1.
\end{equation}

Thus, the ARQ captures how much more expensive it is to execute the Cross-Trace Verification Protocol (CTVP) compared to a single inference. Example token-FLOP estimates used in our experiments are:
\[
\mathrm{FLOPS}_{\text{token}}(7\mathrm{B}) = 1.4\times 10^{10}, \qquad
\mathrm{FLOPS}_{\text{token}}(1.5\mathrm{B}) = 3.0\times 10^{9}.
\]

The implementation also enforces minimum runtime thresholds to prevent degenerate estimates.

\section*{Information-Theoretic Bounds}
Let \(\mathcal{M}\) be the trace space of size \(M=|\mathcal{M}|\) and \(\mathcal{N}\) the program-space of size \(N=|\mathcal{N}|\). Under a conservative independence assumption, the probability that a backdoored model (or an accidental adversary) is \emph{consistent} across \(k\) independently sampled semantic variants is
\begin{equation}\label{eq:prob-consistent}
\Pr\bigl(\text{consistent}\mid\text{backdoor}\bigr) \;=\; \Bigl(\frac{N}{M}\Bigr)^{\,k-1}.
\end{equation}
Hence the expected number of adversarial trials required to find a backdoor that is consistent across \(k\) samples is
\begin{equation}\label{eq:expected-trials}
\mathbb{E}[\text{trials}] \;=\; \frac{1}{\Pr(\text{consistent}\mid\text{backdoor})}
\;=\; \Bigl(\frac{M}{N}\Bigr)^{\,k-1}.
\end{equation}
We approximate typical sizes by
\[
M \approx \bigl(n_{\text{vars}}\cdot V\bigr)^{\,n_{\text{steps}}},\qquad
N \approx n_{\text{vars}}!\cdot n_{\text{blocks}}!,
\]
where \(V\) is the effective per-variable value range (we use \(V\!=\!1000\) in our estimates). In practice \(M \gg N\), so the cost in \eqref{eq:expected-trials} grows exponentially in \(k\).

Equivalently, using trace entropy \(H(\mathcal{T})=\log_2 M\), the expected trials scale as
\begin{equation}
\mathbb{E}[\text{trials}] \approx 2^{(k-1)\,H(\mathcal{T})}.
\end{equation}
These bounds formalize the \emph{non-gamifiability} intuition: an adversary must search an exponentially large space (in \(k\)) to construct a backdoor that remains consistent across a sampled orbit, making such attacks computationally infeasible under realistic assumptions.

We can further derive a theoretical lower bound for ARQ using information-theoretic reasoning.  
If each new trace adds $H$ bits of entropy, then the number of configurations required to distinguish adversarial behavior grows exponentially:
\begin{equation}
\text{Config}_{\min} \approx 2^{kH}.
\end{equation}
In implementation, the function \texttt{calculate\_theoretical\_lower\_bound} uses:
\begin{equation}
2^{(k \cdot H)}, \quad \text{with} \quad H \approx \log_2(N_{\mathrm{samples}}),
\end{equation}
where $N_{\mathrm{samples}}$ is the number of observed trace samples.  
This emphasizes that verification complexity scales exponentially with orbit size $k$, reinforcing the necessity of bounded, probabilistic verification rather than exhaustive exploration.

\section*{Decision logic}
Given a threshold $\tau\in[0,1]$, the final decision rule is:
\[
\text{flag}(P) \;=\; 
\begin{cases}
\text{``ALERT''}, & \mathrm{ARQ}(P,M,\mathcal{O}) < \tau,\\[4pt]
\text{``OK''}, & \mathrm{ARQ}(P,M,\mathcal{O}) \ge \tau.
\end{cases}
\]
Thresholds may be set via calibration on benign and known-adversarial examples.

\paragraph{Caching and reuse.}
An orbit $\mathcal{O}$ may be cached and reused if the identical $(P,k)$ pair reappears; when reusing a cached orbit include a timestamp or a transformation-seed tag to ensure reproducibility of experiments.

\paragraph{Remarks.}
\begin{itemize}
  \item The choice of transformations, the minimum distance ratio $\delta_{\min}$, the attempt multiplier $m_{\max}$, and the failure threshold $r_{\max}$ are hyperparameters that trade off orbit diversity, semantic fidelity, and generation cost.
  \item The similarity function $\operatorname{sim}$ should be chosen to reflect the threat model (e.g., control-flow vs. data-flow similarities).
  \item The consistency score $C$ in \eqref{eq:consistency} is intentionally modular: you can replace it with a robust statistic (trimmed mean, median) or a weighted average to emphasize important trace regions, depending on the application's robustness requirements.
\end{itemize}

\section*{Trace Prediction via Language Model}

For each program $P$ (original or transformed), we predict its execution trace using a large language model (LLM) as an oracle. The model is prompted in natural language, e.g.,
\emph{``Simulate executing this Python program step-by-step. Your trace:''} followed by the code of $P$.  
The textual response is parsed as an execution trace
\[
\tau = [s_1, s_2, \dots],
\]
where each element $s_i$ represents a partial program state (variable assignments and outputs).  
Formally, at each step $i$, we record a mapping of the form $\{x \mapsto v\}$ for each variable $x$ and its corresponding value $v$.

The predictor caches traces by hashing the program string to avoid redundant LLM calls.  
If prediction fails (e.g., parsing error or timeout), an invalid trace is returned.

Let $\tau(P)$ denote the execution trace returned by the model as an \texttt{ExecutionTrace} object.  
Before inference, the model and tokenizer are loaded with robust fallback strategies. Prompt construction is deterministic, and inference uses configuration-defined sampling parameters (e.g., temperature, top-$p$).  
Although no closed-form equation exists, conceptually we view this as a probabilistic mapping:
\begin{equation}
\tau(P) = \mathrm{LLM}\big(\mathrm{prompt}(P)\big),
\end{equation}
where the prompt requests a step-by-step simulation.  
Since LLM inference is stochastic, even semantically equivalent programs may yield differing traces.

\subsection*{Trace Similarity Metric}

To compare two traces $\tau_1$ and $\tau_2$, we compute a weighted similarity
\[
s(\tau_1, \tau_2) \in [0,1].
\]
Let $n_1$ and $n_2$ be the number of steps (excluding the final output) in each trace.  
We define three components: step-length, state, and output similarity.

\paragraph{Step-length similarity.}
\begin{equation}
S_{\mathrm{len}} = \frac{\min(n_1, n_2)}{\max(n_1, n_2)}.
\end{equation}

\paragraph{State similarity.}
Let $m$ be the count of steps $i = 1, \dots, \min(n_1, n_2)$ for which the variable assignments in $\tau_1$ and $\tau_2$ at step $i$ are identical. Then
\begin{equation}
S_{\mathrm{state}} = \frac{m}{\min(n_1, n_2)}.
\end{equation}

\paragraph{Output similarity.}
We define the final-output agreement as
\[
S_{\mathrm{out}} =
\begin{cases}
1, & \text{if final outputs are identical},\\[4pt]
0.5, & \text{if one output is missing},\\[4pt]
0, & \text{otherwise.}
\end{cases}
\]

\paragraph{Overall similarity.}
The combined similarity score is a convex combination:
\begin{equation}\label{eq:trace-sim}
s(\tau_1, \tau_2)
= \alpha\,S_{\mathrm{len}} + \beta\,S_{\mathrm{state}} + \gamma\,S_{\mathrm{out}},
\end{equation}
where $\alpha, \beta, \gamma \in [0,1]$ are configuration weights satisfying
\begin{equation}
\alpha + \beta + \gamma = 1.
\end{equation}
The implementation bounds $s(\tau_1,\tau_2)$ to $[0,1]$ explicitly.  
In prose, this metric compares (i) how similar the lengths of traces are, (ii) how many intermediate states match exactly, and (iii) whether final outputs coincide, combining these aspects into a unified score.


\section*{Cross-Trace Verification Protocol (CTVP)}

Using the semantic orbit and the similarity metric, we perform phase-wise verification on the input program~$P$.  
Let $\tau_0 = \tau(P)$ denote the execution trace of the original program under the \emph{untrusted} predictor, and let $\{Q_1, \dots, Q_k\}$ be the generated orbit.  
For each variant $Q_i$, we obtain the corresponding trace $\tau_i = \tau(Q_i)$ and collect all valid traces:
\[
\mathcal{T} = \{\tau_i : i = 0,1,\dots,k\} \quad (\tau_0 \text{ included}),
\]
ignoring invalid or error traces.

We now compute a \emph{consistency score} among these traces by evaluating all pairwise similarities
$s(\tau_i,\tau_j)$ for $i < j$ and $\tau_i,\tau_j \in \mathcal{T}$.  
Let $\{s_{ij}\}$ denote this multiset of similarity values.  
To ensure robustness to outliers, we take a high percentile (e.g., $p=80$) of these values:
\begin{equation}
C = \mathrm{percentile}_p\big(\{s_{ij}\}\big).
\end{equation}
Here $C \in [0,1]$ represents the cross-trace consistency score—high $C$ (close to~1) implies strong agreement among traces, whereas low $C$ indicates divergent behavior.

\paragraph{Decision logic.}
Let $T_{\mathrm{strict}}$ and $T_{\mathrm{soft}}$ denote the strict and soft thresholds, respectively.  
The final decision is:
\[
\text{Decision} =
\begin{cases}
\text{REJECT}, & C < T_{\mathrm{strict}},\\[4pt]
\text{Check trusted model}, & T_{\mathrm{strict}} \le C < T_{\mathrm{soft}},\\[4pt]
\text{ACCEPT}, & C \ge T_{\mathrm{soft}}.
\end{cases}
\]

\paragraph{Trusted model verification.}
In the intermediate regime ($T_{\mathrm{strict}} \le C < T_{\mathrm{soft}}$), we optionally engage a \emph{trusted} model for verification.  
Let $\tau_{\mathrm{trusted}} = \tau_{\mathrm{trusted}}(P)$ be the trusted trace, and compute
\begin{equation}
s_{\mathrm{trusted}} = \mathrm{median}_i\big[s(\tau_{\mathrm{trusted}}, \tau_i)\big],
\end{equation}
where $\tau_i \in \mathcal{T}$.  
We also denote $s_{\mathrm{untrusted}} = C$.  
If either
\[
s_{\mathrm{trusted}} < T_{\mathrm{trust}}
\quad \text{or} \quad
s_{\mathrm{trusted}} - s_{\mathrm{untrusted}} > \Delta,
\]
we label the sample as \textbf{REJECT}; otherwise, \textbf{ACCEPT}.  
In practice, $\Delta \approx 0.15$ serves as a divergence threshold.

\begin{algorithm}[t]
\caption{Cross-Trace Verification Protocol (CTVP)}
\label{alg:ctvp}
\begin{algorithmic}[1]
\Require Program $P$, orbit size $k$, thresholds $T_{\mathrm{strict}}, T_{\mathrm{soft}}, T_{\mathrm{trust}}$
\Ensure Verdict $\in \{\mathrm{ACCEPT}, \mathrm{REJECT}\}$ and diagnostic scores
\State $\tau_0 \gets$ \textsc{PredictTrace}$(P)$ using untrusted model
\If{$\tau_0$ is invalid} \Return \textbf{error}
\EndIf
\State $\{Q_i\}_{i=1}^k \gets$ \textsc{GenerateOrbit}$(P, k)$
\State $\mathcal{T} \gets \{\tau_0\}$
\For{each $Q_i$}
    \State $\tau_i \gets$ \textsc{PredictTrace}$(Q_i)$
    \If{$\tau_i$ is valid}
        \State add $\tau_i$ to $\mathcal{T}$
    \EndIf
\EndFor
\If{$|\mathcal{T}| < 2$} \Return \textbf{REJECT} (insufficient traces)
\EndIf
\State Compute pairwise similarities $s_{ij} = s(\tau_i, \tau_j)$ for $\tau_i, \tau_j \in \mathcal{T}$
\State $C \gets \mathrm{percentile}_p(\{s_{ij}\})$ \Comment{Consistency score}
\If{$C < T_{\mathrm{strict}}$}
    \Return \textbf{REJECT}
\ElsIf{$C < T_{\mathrm{soft}}$}
    \State $\tau_{\mathrm{trusted}} \gets$ \textsc{PredictTrace}$(P)$ using trusted model
    \If{$\tau_{\mathrm{trusted}}$ invalid} \Return \textbf{REJECT}
    \EndIf
    \State $s_{\mathrm{trusted}} \gets \mathrm{median}_{\tau \in \mathcal{T}} s(\tau_{\mathrm{trusted}}, \tau)$
    \State $g \gets s_{\mathrm{trusted}} - C$
    \If{$s_{\mathrm{trusted}} < T_{\mathrm{trust}}$ \textbf{or} $g > 0.15$}
        \Return \textbf{REJECT}
    \Else
        \Return \textbf{ACCEPT}
    \EndIf
\Else
    \Return \textbf{ACCEPT} \Comment{Optionally enable runtime monitoring}
\EndIf
\end{algorithmic}
\end{algorithm}

This protocol enforces that all semantically equivalent variants produce consistent traces under the untrusted model.  
A low consistency score~$C$ indicates potential hidden or malicious behavior.  
In the ``soft'' regime, trusted-model verification reduces false positives: if both untrusted and trusted traces disagree significantly, the program is rejected; otherwise, it is accepted with confidence.

\section*{Theoretical Space Estimation}

We quantify the theoretical search spaces of program transformations and execution traces to justify the difficulty of adversarially manipulating these features.  
A \texttt{TheoreticalSpaceEstimator} computes two quantities:

\begin{itemize}
    \item \textbf{Trace space size} $M$: the number of distinct execution traces.
    \item \textbf{Program space size} $N$: the number of distinct programs reachable via syntactic transformations.
\end{itemize}

Let $n_{\mathrm{id}}$ denote the number of unique identifiers (variables or functions) in program $P$, and $n_{\mathrm{blk}}$ the number of top-level code blocks.  
Let $L$ be the number of executable statements (approximate computational steps), and $R$ be the average range of variable values (a hyperparameter from configuration).  
Then, the theoretical estimator uses the following approximations:
\begin{align}
M &\approx (n_{\mathrm{id}} \cdot R)^{L}, \label{eq:trace_space}\\[4pt]
N &\approx n_{\mathrm{id}}! \times n_{\mathrm{blk}}!. \label{eq:program_space}
\end{align}

Equation~\eqref{eq:trace_space} estimates the trace-space size $M$ by assuming each of $n_{\mathrm{id}}$ variables can take approximately $R$ distinct values at each of $L$ execution steps.  
Equation~\eqref{eq:program_space} estimates the program-space size $N$ as the number of possible renamings of identifiers and permutations of independent blocks.  
In practice, extremely large values are capped numerically for stability.

These estimates demonstrate that even small programs have astronomically large semantic and syntactic variation spaces. This motivates the need for \emph{sampling} (via the semantic orbit) rather than exhaustive enumeration.


\subsection*{Baseline Detection Methods}

For comparison, we implement simple static baselines that do not rely on the LLM but instead check for obvious malicious patterns or anomalies.

\paragraph{Static code patterns.}
We flag a program $P$ if its source contains suspicious tokens such as \texttt{eval(}, \texttt{exec()}, secret constants, imports of \texttt{os.system}, or keywords like \texttt{BACKDOOR}.
Formally, this corresponds to a boolean predicate
\begin{equation}
b_{\mathrm{static}}(P) \in \{0,1\},
\end{equation}
which evaluates to $1$ if any predefined regular expression pattern matches the program text:
\[
b_{\mathrm{static}}(P) = 
\begin{cases}
1, & \text{if }\exists\, \text{pattern } r \text{ s.t. } r \text{ matches } P,\\[4pt]
0, & \text{otherwise.}
\end{cases}
\]

\paragraph{Complexity heuristic.}
We parse $P$ into an abstract syntax tree (AST) and compute:
\[
n = \text{total number of AST nodes}, \quad 
c = \text{number of control nodes (if, while, for, try, etc.)}.
\]
If the code is either overly complex or underspecified, we flag it as suspicious. Specifically:
\begin{equation}
b_{\mathrm{complex}}(P) =
\begin{cases}
1, & n > 50 \text{ and } \tfrac{c}{n} > 0.3,\\[4pt]
1, & n < 20 \text{ and } c > 3,\\[4pt]
0, & \text{otherwise.}
\end{cases}
\label{eq:complex}
\end{equation}

\paragraph{Entropy heuristic.}
We compute the Shannon entropy $H$ of the character distribution in $P$:
\begin{equation}
H = - \sum_{x \in \Sigma} \frac{n_x}{N} \log_2 \frac{n_x}{N},
\end{equation}
where $n_x$ is the count of character $x$, $N = |P|$ is the total number of characters, and $\Sigma$ is the program’s character alphabet.
Programs with extremely low or high entropy are flagged as potentially malicious:
\begin{equation}
b_{\mathrm{entropy}}(P) =
\begin{cases}
1, & H < 3.0 \text{ or } H > 6.0,\\[4pt]
0, & \text{otherwise.}
\end{cases}
\label{eq:entropy}
\end{equation}

\paragraph{Weighted combination.}
The final baseline flags $P$ as malicious if at least two of the above three checks trigger. Let
\[
b_{\mathrm{static}},\; b_{\mathrm{complex}},\; b_{\mathrm{entropy}} \in \{0,1\}.
\]
Then
\begin{equation}
b_{\mathrm{weighted}}(P) =
\mathbb{I}\!\left[(b_{\mathrm{static}} + b_{\mathrm{complex}} + b_{\mathrm{entropy}}) \ge 2\right],
\label{eq:weighted}
\end{equation}
where $\mathbb{I}[\cdot]$ denotes the indicator function.

\section{Malware Detection Performance Analysis}

\begin{table}[h!]
\centering
\caption{Baseline Detection Methods Performance}
\label{tab:baseline_performance}
\begin{tabular}{lccc}
\toprule
\textbf{Method} & \textbf{Detection Rate (\%)} & \textbf{False Positive Rate (\%)} & \textbf{Benign Acceptance (\%)} \\
\midrule
Static Analysis & 51.7 & 0.0 & 100.0 \\
Complexity & 80.0 & 62.5 & 37.5 \\
Entropy & 0.0 & 0.0 & 100.0 \\
Weighted & 51.7 & 0.0 & 100.0 \\
CTVP & 83.8 & 98.8 & 1.2 \\
\bottomrule
\end{tabular}
\end{table}

\begin{table}[h!]
\centering
\caption{Information-Theoretic Analysis by Orbit Size}
\label{tab:info_theoretic}
\resizebox{\textwidth}{!}{
\begin{tabular}{c c c c c c c c}
\toprule
\textbf{Orbit Size} &
\textbf{Transformation Success Rate (\%)} &
\textbf{Generated Variants} &
\textbf{Avg.\ Traces / Program} &
\textbf{Malicious Detection} &
\textbf{Benign False Positives} &
\textbf{Detection Rate (\%)} &
\textbf{FPR (\%)} \\
\midrule
3  & 60.0 & 180 / 300   & 1.60 & 49 / 60 & 40 / 40 & 81.7 & 100.0 \\
5  & 77.6 & 388 / 500   & 2.80 & 50 / 60 & 39 / 40 & 83.3 & 97.5  \\
10 & 79.4 & 794 / 1000  & 4.30 & 50 / 60 & 39 / 40 & 83.3 & 97.5  \\
20 & 78.6 & 1572 / 2000 & 7.60 & 52 / 60 & 40 / 40 & 86.7 & 100.0 \\
\bottomrule
\end{tabular}
}
\end{table}

\begin{table}[h!]
\centering
\caption{ARQ Ratio Statistics by Orbit Size}
\label{tab:arq_statistics}
\begin{tabular}{ccccc}
\toprule
\textbf{Orbit Size} & \textbf{Mean} & \textbf{Median} & \textbf{Min} & \textbf{Max} \\
\midrule
3 & 4.00 & 4.00 & 4.00 & 4.00 \\
5 & 6.00 & 6.00 & 6.00 & 6.00 \\
10 & 11.00 & 11.00 & 11.00 & 11.00 \\
20 & 21.00 & 21.00 & 21.00 & 21.00 \\
\bottomrule
\end{tabular}
\end{table}

\begin{table}[h!]
\centering
\caption{Consistency Score Statistics by Orbit Size}
\label{tab:consistency_scores}
\begin{tabular}{cccc}
\toprule
\textbf{Orbit Size} & \textbf{Mean} & \textbf{Median} & \textbf{Standard Deviation} \\
\midrule
3 & 0.881 & 0.925 & 0.125 \\
5 & 0.835 & 0.821 & 0.113 \\
10 & 0.818 & 0.802 & 0.126 \\
20 & 0.830 & 0.790 & 0.107 \\
\bottomrule
\end{tabular}
\end{table}

\paragraph{Discussion.}
These baselines provide interpretable points of comparison for our detection performance but are generally weaker than the dynamic cross-trace method described earlier, as they cannot capture semantic inconsistencies or adversarial obfuscation.

\section{Open Research Questions}

\begin{tcolorbox}[myboxstyle]
\begin{itemize}
    \item How can we reduce false positive rates below 10\% while maintaining detection rates above 80\%?
    \item What similarity metrics and threshold calibration methods achieve optimal precision-recall trade-offs across diverse program domains?
    \item Can hybrid approaches combining LLM-based trace prediction with sandboxed execution improve both accuracy and efficiency?
    \item What is the relationship between transformation diversity, orbit size, and detection coverage for different backdoor types?
    \item Can CTVP's semantic orbit analysis generalize to other code generation tasks beyond Python (e.g., C++, JavaScript, SQL)?
    \item How do adaptive adversaries respond to CTVP, and what countermeasures can maintain robustness against co-adaptive attacks?
\end{itemize}
\end{tcolorbox}

\end{document}